\documentclass[11pt,a4paper]{article}
\usepackage{authblk} 
\usepackage[hyperref]{acl2021}
\usepackage{times}
\usepackage{latexsym}
\usepackage{graphicx}
\usepackage{amsmath}
\usepackage{multirow}
\usepackage{url}

\usepackage{microtype}

\aclfinalcopy 

\title{ArgFuse: A Weakly-Supervised Framework for Document-Level Event Argument Aggregation}

\author[1]{\textbf{Debanjana Kar}}
\author[2]{\textbf{Sudeshna Sarkar}}
\author[3]{\textbf{Pawan Goyal}}
\affil[ ]{Department of Computer Science \& Engineering}
\affil[ ]{Indian Institute of Technology, Kharagpur.}
\affil[1]{debanjana.kar@iitkgp.ac.in}
\affil[2,3]{\textit {\{sudeshna, pawang\}@cse.iitkgp.ac.in}}

\date{}

\begin{document}
\maketitle
\begin{abstract}
Most of the existing information extraction frameworks \cite{wadden2019entity, veyseh2020graph} focus on  sentence-level tasks and are hardly able to capture the consolidated information from a given document. In our endeavour to generate precise document-level information frames from lengthy textual records, we introduce the task of Information Aggregation or Argument Aggregation. More specifically, our aim is to filter irrelevant and redundant argument mentions that were extracted at a sentence level and render a document level information frame.
Majority of the existing works have been observed to resolve related tasks of document-level event argument extraction \cite{yang-etal-2018-dcfee, zheng-etal-2019-doc2edag} and salient entity identification \cite{jain-etal-2020-scirex} using supervised techniques. To remove dependency from large amounts of labeled data 
, we explore the task of information aggregation using weakly-supervised techniques. In particular, we present an extractive algorithm with multiple sieves which adopts active learning strategies to work efficiently in low-resource settings.
For this task, we have annotated our own test dataset comprising of 131 document information frames and have released the code and dataset to further research prospects in this new domain. To the best of our knowledge, we are the first to establish baseline results for this task in English. Our data and code are publicly available at \url{https://github.com/DebanjanaKar/ArgFuse}
\end{abstract}

\section{Introduction}
Extraction of event argument information at a document level is an important non-trivial task that requires a system to have advanced natural language understanding capabilities. Most of the existing event-argument extraction systems \cite{ nguyen2016joint, luan2019general, wadden2019entity, veyseh2020graph} pertain to a sentence-level focus, often circumventing to capture information at a document-level. Among the few existing works that have researched the task of document-level event argument extraction, we observe that unsupervised techniques like \cite{Hamborg2019b} lack the capacity to identify complex argument mentions, while sophisticated supervised mechanisms like that of \cite{yang2018dcfee, zheng2019doc2edag} rely on large amounts of annotated corpus and present domain-specific solutions.\\ 
\begin{figure}
    \centering
    \includegraphics[scale=0.43]{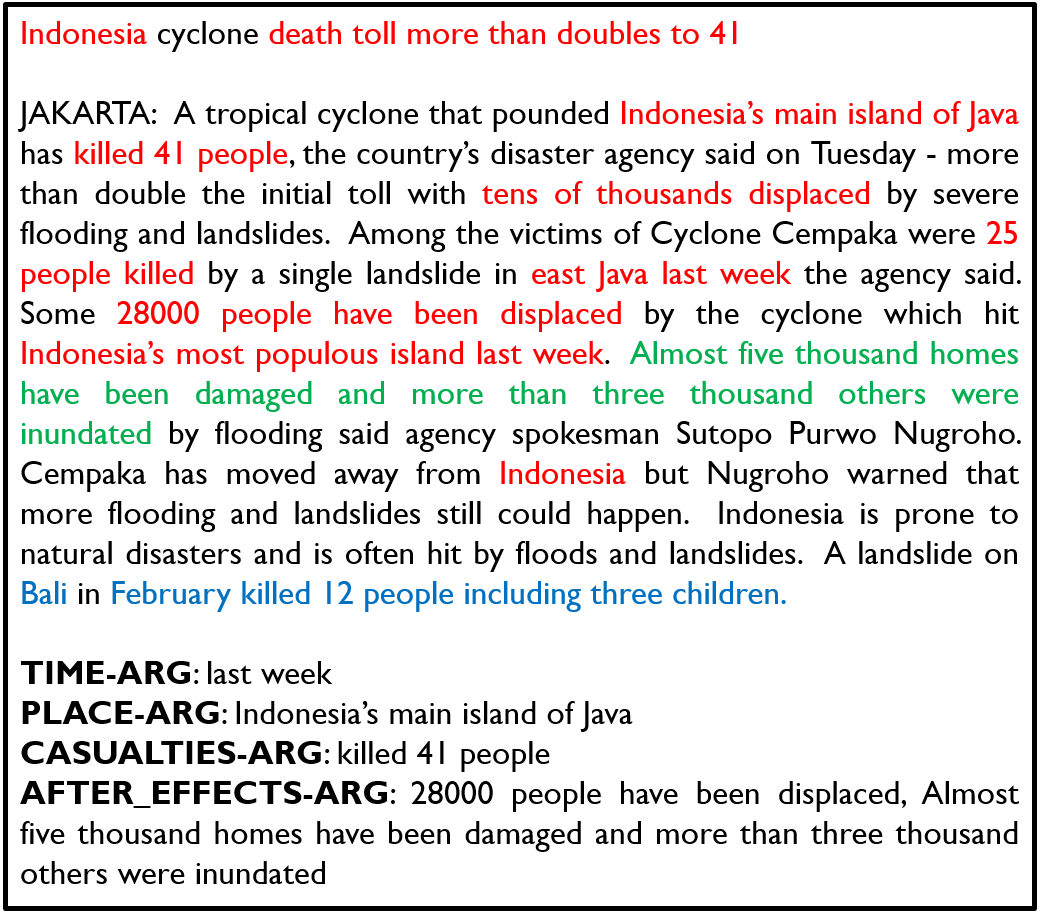}
    \caption{Example document excerpt from our corpus highlighting the different challenges of the aggregation task. The phrases highlighted in red, blue and green denote the redundant, irrelevant and exclusive sentence-level arguments respectively. The document level arguments are as reported at the end of the example.}
    \label{fig:eg1}
\end{figure}
While supervised techniques may often produce highly accurate systems, in real life, annotating a large corpus can be both very expensive and time consuming. In order to surmount the existing shortcomings coupled with the challenging scenario of data scarcity, we propose our model \textit{ArgFuse}. 
ArgFuse focuses on extraction of relevant and non-redundant event arguments at a document level scope. 
The task of document level event argument extraction typically focuses on extracting argument mentions associated with an event type or event trigger from a document. It does not involve checking extracted arguments for irrelevant and redundant mentions. Through our work, we also propose the related task of Argument Aggregation which focuses on assessing extracted arguments for irrelevance and redundancy to produce a precise aggregated document-level information frame.
Figure \ref{fig:eg1} provides an illustrative example of our task while highlighting the different challenges the task presents.\\
The task presents itself with multiple challenges and avenues to explore. 
To produce a precise document-level information frame, we focus on sieving irrelevant and redundant sentence-level argument mentions. Irrelevant argument mentions, as illustrated in Figure \ref{fig:eg1} refer to argument mentions that do not contribute to the topical focus of the document. A category of irrelevant arguments often encountered in real life news articles are past event records mentioned in narratives to provide a comparative perspective to the reader, much like the arguments in blue mentioned in Figure \ref{fig:eg1}. While irrelevant arguments are usually mentions that refer to past, future or unrelated events; redundancy in arguments manifests in diverse forms. Redundant arguments can either be i) duplicate argument mentions (e.g. \textit{last week}), ii) arguments with similar surface form (e.g. \textit{Indonesia, Indonesia's main Island Java)}, iii) re-worded (e.g. \textit{killed 41 people, death toll more than doubles to 41}), or iv) subsuming information of the other argument(s) (e.g. \textit{killed 41 people, 25 people killed}). While the first two types of redundancy can be tackled by simple heuristics, the detection of the remaining types of redundancy requires implicit natural language understanding and coreference reasoning capabilities. To filter such arguments effectively, we realised that contextual information of a document is imperative. The argument mentions in green illustrated in Figure \ref{fig:eg1} highlight arguments which impart unique information with respect to the context of the document. These argument mentions cannot be aggregated and are directly added to the output information frame. We refer to such argument mentions in our work as \textit{exclusive argument} mentions.\\
Based on the different types of arguments we encounter, we propose an extractive algorithm that aggregates sentence-level argument or entity mentions to produce precise document-level information frames from lengthy text articles effectively. In our work, we present an end-to-end framework to extract events and arguments from English news articles and present an aggregated information frame at the document level. Given that we introduce a novel task with no prior labeled dataset, we present a weakly-supervised algorithm to achieve our task with a good accuracy. Our contributions in this work are two fold:
\begin{enumerate}
    \item We propose a novel task of document-level event-argument aggregation and establish baseline for the same. We also release the first annotated test dataset for this task with 131 aggregated document information frames.
    \item We propose a weakly supervised model to aggregate event-arguments at a document-level. We deeply analyse the task, dataset and algorithm proposed in this paper, thus highlighting areas of future research and development.
\end{enumerate}
In the following sections, we discuss our dataset, algorithms and findings in detail. Our analysis emphasizes on the importance of having document level information extraction frameworks for the task of argument aggregation and we invite the research community to further investigate this task.
\section{Related Work}
Event argument extraction is a well researched information extraction task which has seen a lot of work at the sentence-level \cite{wang-etal-2019-hmeae,wadden2019entity,nguyen2016joint, luan-etal-2019-general,veyseh2020graph} but a scarce amount of research has been carried out at the document level. Recent literature on event argument extraction at a document level include the works of \cite{yang-etal-2018-dcfee} and \cite{zheng-etal-2019-doc2edag}. While  \cite{zheng-etal-2019-doc2edag}'s work explores supervised transformer based techniques to extract events and arguments from Chinese financial documents, \cite{yang-etal-2018-dcfee} employs Bi-LSTM based classifiers on a subset of the same dataset to extract events first at the sentence level followed by a document level extraction similar to our framework. \cite{jain-etal-2020-scirex} employs a BiLSTM-CRF classifier to finetune SciBERT\cite{beltagy2019scibert} on various document-level information extraction tasks including the related task of salient entity clustering. All of these methods employ supervised techniques which call for a large corpus of annotated dataset, making it difficult to adopt to domains and tasks with no labelled annotations. \cite{Hamborg2019b} presents an unsupervised approach of extracting document level information from news documents, but the heuristics adopted in their work do not extend well to our task which involves more complex argument mentions. The limited amount of research that exists in this domain does not explore the task of aggregation in particular, where, given a set of arguments referring to the same concept, the most informative argument is selected to represent that knowledge. We propose a novel task and present an end-to-end baseline solution to extract and aggregate document-level arguments which presents a complete overview of the document without minimal loss of information. In our work, we employ ranking strategies as part of our aggregation process. Some of the classic works related to the task of ranking text snippets are that of PageRank \cite{Pageetal98} and TextRank \cite{mihalcea-tarau-2004-textrank}.
\section{Dataset}
One of the main challenges that we faced was the unavailability of annotated resources for this task. For our auxiliary task of sentence level event argument extraction, we use the dataset adopted by \cite{kar2021event}. The dataset is available in five Indian languages but for this task, we only use the English dataset. The dataset covers 32 event types at a fine grain level and 12 event types at a coarse level. The dataset contains annotations for 14 argument types, but in our work we focus on 6 main argument types which are, \textit{Time, Place, Casualties,  After-Effect, Reason} and \textit{Participant}. In the sections to follow we discuss regarding the scarcely available document-level annotated resources and the details of the annotated dataset we release with this work.
\subsection{Existing Document-Level IE Datasets}
Information Extraction (IE) is a well-researched domain albeit mostly at the sentence-level. Event-Argument Extraction, the IE task most related to the task of aggregation has a number of well-documented and reliable datasets annotated at the sentence level in different languages like ACE 2005 and TAC KBP \cite{mitamura-etal-2015-event} datasets. IE tasks with a document-level focus have gained attention in recent times but there are hardly any document-level event argument annotated datasets. We discuss two recent works that include document-level event argument or entity mention annotations here; the RAMS \cite{ebner-etal-2020-multi} and the SciREX \cite{jain-etal-2020-scirex} dataset. The RAMS dataset is not particularly document-level, but explores the task of extracting argument roles beyond sentences. In a 5-sentence window of a news article around each event trigger, they annotate the closest span for each argument role. Their ontology consists of 139 event types and 65 argument roles. The SciREX dataset is a comprehensive dataset comprising of document-level annotations for a variety of IE tasks. The dataset consists of annotations for related tasks like entity recognition and coreference on 438 scientific articles. However, none of these datasets provide consolidated argument annotations for an entire document. To the best of our knowledge, we are the first to introduce a document-level event-argument annotated dataset in English which provides an aggregated overview of the document, that is, the first annotated dataset for the task of Event-Argument Aggregation in English.

\subsection{ArgFuse Dataset}
\begin{figure}[ht]
    \centering
    \includegraphics[scale=0.45]{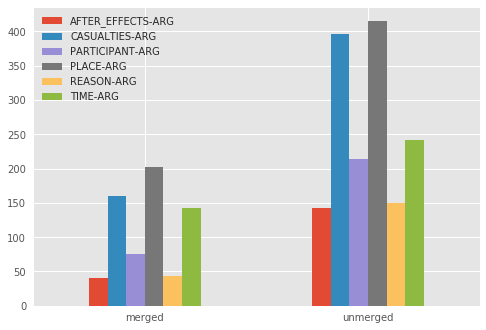}
    \caption{Distribution of sentence-level and document-level argument mentions in the annotated ArgFuse dataset. In the figure, merged refers to document-level annotations and unmerged refers to sentence-level annotations. }
    \label{fig:dataset1}
\end{figure}

\begin{figure}[ht]
    \centering
    \includegraphics[scale=0.45]{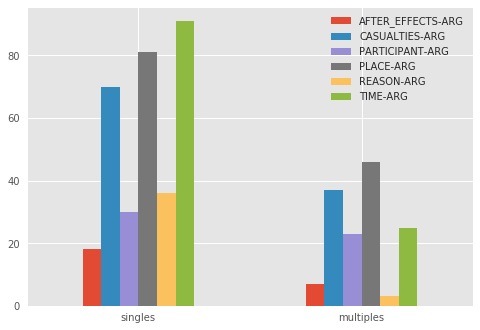}
    \caption{Distribution of single and multiple argument mentions among the document-level arguments in the annotated ArgFuse dataset. \textit{Singles} refer to a  category of argument roles which constitute of a single argument mention and \textit{Multiples} refer to a category of argument roles which constitute of more than one relevant argument mentions at a document-level. }
    \label{fig:dataset2}
\end{figure}

Most of our work employs weakly-supervised techniques for curation of document level information frames but for sound testing of our final model we manually aggregate document-level arguments for each argument type from the 131 English documents in the test set of the above mentioned dataset. Each information frame for a document contains the event-type and the corresponding relevant arguments for each argument type from the document. During curation of the test set, we followed certain annotation guidelines defined for each argument type. 
The guidelines contain detailed instructions for identifying relevant arguments at a document level. For example, if the \textit{Time} arguments of a document mention different degrees of temporal expressions like day, month and hour of the day, all the arguments are to be considered as relevant and aggregation is not required. The dataset was curated by two research scholars with good domain knowledge. We report the statistics of our dataset in Figures \ref{fig:dataset1} and \ref{fig:dataset2}. In figure \ref{fig:dataset1} we can observe the amount of redundancy and irrelevance prevalent in the extracted sentence-level information. In figure \ref{fig:dataset2}, we observe that although a number of argument roles in a document constitute of a single relevant argument mention (referred to as \textit{Singles}), a significant number of argument roles constitute of multiple number of relevant argument mentions (referred to as \textit{Multiples}). This highlights the fact that the number of relevant arguments for a particular argument role or type can be flexible and the model should be able to accommodate that flexibility. We release the manually annotated test set along with the annotation guidelines to further research prospects in this novel task.\footnote{\url{https://github.com/DebanjanaKar/ArgFuse}}


\section{Information Aggregation}
\begin{figure}
    \centering
    \includegraphics[scale=0.38]{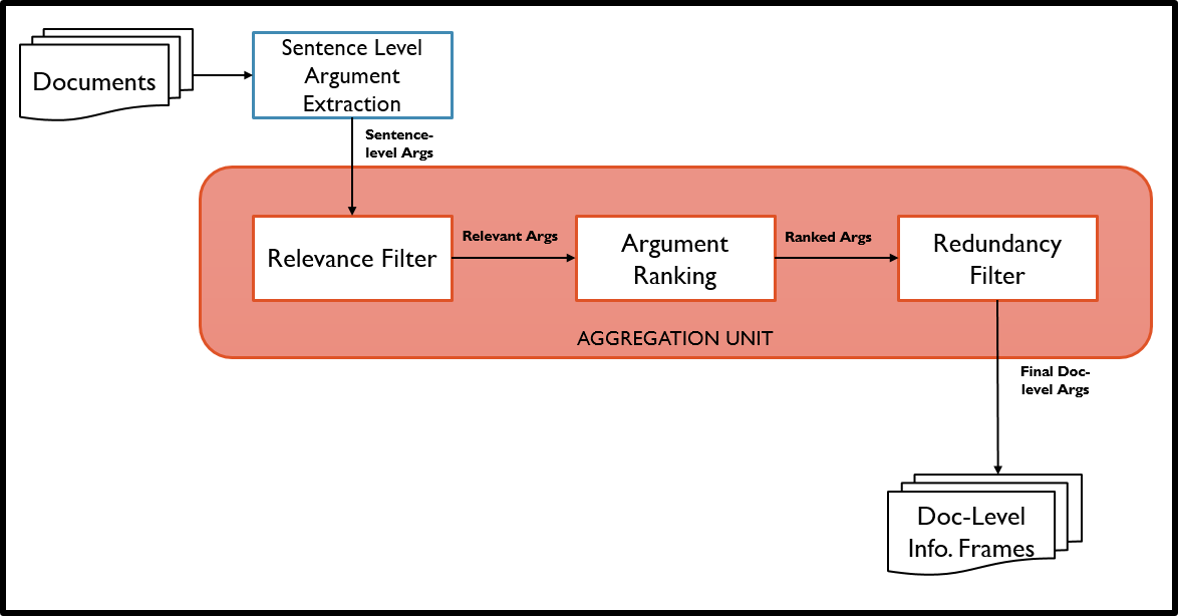}
    \caption{General overview of the complete argument aggregation framework.}
    \label{gen_arch}
\end{figure}

In this section, we detail the approaches that were taken to build a weakly-supervised argument aggregation framework. The framework primarily involves two main modules: i) Sentence-level Information Extraction (IE), which extracts the sentence-level arguments along with their event type, the ii) Aggregation Unit, which renders document-level information frames. A general overview of the pipeline is illustrated in Figure \ref{gen_arch}. We explain each of these modules in detail in the sections to follow.

\subsection{Event Argument Extraction}

Given a document $D$ of event type $E$, the objective of this sub-task is to extract the argument label sequence $(y_{1},..., y_{n})$ for the corresponding word sequence $(x_{1},..., x_{n})$, $n$ being the number of tokens in $D$. For example, for a given document: \textit{``The flood waters destroyed 500 homes in Assam ...''}, the corresponding label sequence would be: {`O O O AFTER\_EFFECTS\_ARG AFTER\_EFFECTS\_ARG AFTER\_EFFECTS\_ARG O PLACE\_ARG ...'}.
To ensure high accuracy and low error propagation, we have adopted \cite{kar2021event}'s approach of sentence-level event argument extraction using causal knowledge structures for this sub-task. \cite{kar2021event}'s approach provides state-of-the-art results on the INDEE dataset \cite{maheshwari2020tale} and ensures efficient extraction of the low resource, complex causal arguments like \textit{Reason} and \textit{After-Effects} using the specially designed causality feature. The causality feature for each event consists of words and phrases which are used frequently in a causal context for particular event scenarios. The input document, concatenated with the feature at either extremes is encoded using a fine-tuned BERT encoder and each token is ultimately classified to one of the six argument types in the TO format (adopted from \cite{maheshwari2020tale}). While we adopt \cite{kar2021event}'s approach in our work, the event extraction module can be easily substituted with more suitable models in the future thus leveraging the modular nature of our algorithm. Using \cite{kar2021event}'s approach, we extract the sentence-level event arguments for our corpus with an F1 score of $86.12 \%$.

\subsection{Aggregation}
\begin{figure*}
    \centering
    \includegraphics[scale=0.55]{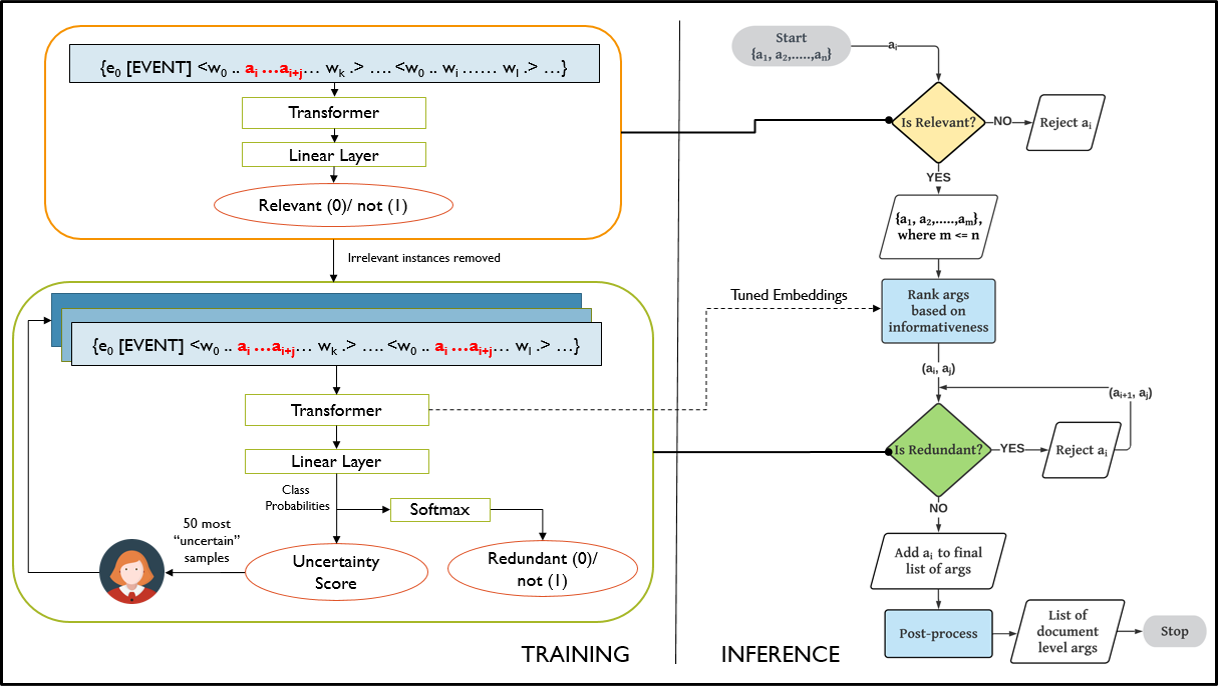}
    \caption{Detailed illustration of ArgFuse. This figure depicts the flow of control during inference with ArgFuse. The spans $(a_i..a_i+j)$ in the relevance and redundancy filters refer to argument mention spans within sentences in the input document.}
    \label{fig:my_label}
\end{figure*}

The aggregation unit is the primary module which identifies the most relevant and informative arguments from a pool of sentence-level argument mentions at a document-level scope. As illustrated in Figure \ref{gen_arch}, the aggregation unit consists of i) the Relevance Filter, which sieves the irrelevant arguments, ii) the Argument Ranking module, which ranks the arguments based on their informativeness and iii) the Redundancy Filter, which sieves the redundant arguments at a document level. The detailed architecture of the aggregation unit is illustrated in Figure \ref{fig:my_label}.

\subsubsection{Relevance Filter}
Sentence-level IE outputs have often been observed to contain arguments that are not relevant to the document's focus. The main task of the Relevance Filter unit is to sieve such arguments. Given the extracted sentence level arguments of a particular argument type along with the context of its constituent document, the relevance filter proceeds to classify each argument mention as relevant or not. Since we did not possess labelled samples for this task, we manually annotated 500 training and 100 test instances. We observe that identifying irrelevant instances is relatively easier with explicit contextual and syntactic cues. Hence, on fine-tuning a ROBERTA-based classifier for our subtask,even with a very limited number of training instances, we obtain an F1 score of $85 \%$. The performance metrics of the relevance filter is detailed in Table \ref{tab:module-wise}.
\subsubsection{Redundancy Filter}
Detection of redundant arguments is comparatively more challenging compared to the sub-task of relevance detection. While certain groups of argument mentions are explicitly redundant (eg. duplicate mentions, substrings), other groups of redundant argument mentions are more implicit in nature. To effectively identify redundant arguments in a low-resource setting, we employ active learning strategies. Given a pair of arguments $(a_i, a_j)$ along with the context of it's constituent document, we train a binary classifier to maximize $P(y|a_i, a_j)$, where $y = 1$, if $(a_i, a_j)$ are redundant else $y = 0$. To train a binary classifier for such a task, a large annotated corpus would have been effective but in the absence of such a corpus, we adopt the effective technique of active learning with $1045$ manually annotated seed instances. After each epoch of active learning, $50$ most uncertain samples are identified using the Monte Carlo estimation of error reduction  \cite{roy2001toward}. These samples are then manually annotated and transferred from the pool of unannotated test samples to the list of annotated training instances. This process is repeated until we do not see any further improvement in the F1 Score. Based on the findings reported in \cite{hu2018active}, to avoid bias from the previous epoch, we fine-tune the pre-trained BERT-based classifier on the entire annotated dataset for every run of active learning. Once we have the necessary annotations derived from the multi-epoch active learning session, we train our binary classifier using all the annotated samples for $15$ epochs. Our relevance filter is evaluated in Table \ref{tab:module-wise}.
\subsection{Inference}
The steps followed during inference are illustrated in Figure \ref{fig:my_label}. The process starts by employing the trained Relevance filter to segregate relevant argument mentions from irrelevant ones. The relevant argument mentions are then ranked based on their informativeness. Given a list of argument-mentions $(arg_{1}, arg_{2}, ..., arg_{m}); m$ being the total number of mentions in the list of type $t$, our objective is to rank the mentions based on which argument instance imparts greater knowledge about the document's event. To compare the informativeness of the arguments, we rank the arguments using the unsupervised Biased TextRank \cite{kazemi-etal-2020-biased}. Biased TextRank is formally defined as:
\begin{align*}
   R(V_{i}) = Bias Weight*(1-d) +  d* R^{'}(V_{i}) \\
   R^{'}(V_{i}) = {\sum\limits_{V_{j} \in In(V_{i})} \dfrac{w_{ij}}{\sum\limits_{V_{k} \in Out(V_{j})}w_{jk}} * R(V_{j})}
\end{align*}
where, $R(V_{i})$ is the score assigned to the vertex $V_{i}$, $In(V_{i})$ denotes the incoming and $Out(V_{i})$ denotes the outgoing edges from the vertex $V_{i}$. The damping factor $d$ is set to 0.85.
In our task, each of the arguments in the list correspond to a vertex and the vertices are connected by a weighted edge. The weight of each edge is determined by the cosine similarity between two arguments $(arg_{i}, arg_{j})$. The bias weight is determined by calculating the cosine similarity between the bias text and the argument text. The bias texts comprise of  short text snippets defined for each argument type along with the document and it's event type. Each of the arguments and the bias texts are encoded using the tuned embeddings from the redundancy filter. Higher the score of the argument (or vertex), more informative is the argument. The ranked list contains the arguments sorted in a non-decreasing order based on their obtained Biased TextRank scores.\\
Each argument from the ranked list of argument mentions is compared sequentially with the other arguments using the Redundancy filter. If any pair of arguments are classified redundant, the argument with lower score is discarded and the process is continued. To  reduce loss of information further, we adopt the following rules: 
\begin{enumerate}
    \item If an argument type contains a single argument mention extracted at the sentence level, the argument is added to the document-level information frame directly.
    \item If all the mentions in a list of arguments of argument type $t$ are classified irrelevant or redundant rendering a null set, we add the argument with the highest score from the list to the document-level information frame.
\end{enumerate}
Once all the sentence-level arguments of a document have been processed through the above described modules, a precise document-level information frame is rendered.
\section{Experiments \& Results}
In this section we shall detail the execution details of the experiments and analyse the results obtained.
\subsection{Experiments}
We have experimented with different encodings and classifiers in our work. For the Sentence Level Event-Argument Extraction task we have encoded the text using Huggingface's \cite{wolf-etal-2020-transformers} bert-base-multilingual-cased model pre-trained on 104 languages\footnote{https://huggingface.co/bert-base-multilingual-cased} \cite{devlin-etal-2019-bert}. For encoding text in the relevance filter, the pretrained roberta-base- model\footnote{https://huggingface.co/roberta-base}\cite{liu2019roberta} was used while for the redundancy filter, the pretrained bert-base-model\footnote{https://huggingface.co/bert-base-cased} was used. 
The ROBERTA-based relevance filter was trained for $3$ epochs on $500$ training samples while the redundancy filter was trained for $15$ epochs after retrieving required annotations from $5$ epochs of active learning. The batch size for our experiments was $8$. All our experiments were performed on a Tesla K40-C server.

\subsection{Metrics}
To analyse the generated document-level information frames and report the performance of the designed framework, we have adopted Precision, Recall and F1-Score as the metrics of our choice. To calculate the above mentioned metrics we count TP, FP and FN as follows:
\begin{itemize}
    \item \textit{True Positive (TP):} When the detected argument exists in the true argument list. 
    \item \textit{False Positive (FP): } When detected argument does not exist in the true argument list.
    \item \textit{False Negative (FN): } When argument from true argument list is not among the detected arguments.
\end{itemize}

\begin{table}[]
\begin{tabular}{|c|c|c|c|c|}
\hline
\multirow{2}{*}{Modules} & \multicolumn{3}{c|}{Macro} & Micro \\ \cline{2-5} 
                         & P          & R         & F1        & F1            \\ \hline
Relevance Filter         & 81       & 85      & 83      & 85          \\
Redundancy Filter        & 69       & 71      & 70      & 70          \\ \hline
\end{tabular}
\caption{Module-wise performance measure using F1-scores (in \%). Macro \& Micro denote the averaging scheme adopted for these metrics.}
\label{tab:module-wise}
\end{table}

\begin{table}[t]
\begin{tabular}{|c|c|c|c|}
\hline
\textbf{Models}               & \textbf{P}     & \textbf{R}     & \textbf{F1}    \\ \hline
GiveMe5W1H                    & 25.13          & 25.17          & 25.15          \\
TextRank @ k = 1        & 65.63          & 50.33          & 56.97          \\
TextRank @ k = 2        & 66.59          & 51.18          & 57.87          \\
Biased TRank @ k = 1 & \textbf{70.48} & 56.73          & 62.86          \\
Biased TRank @ k = 2 & 60.48          & \textbf{68.91}          & 64.42          \\
ArgFuse                       & 67.98          & 64.90 & \textbf{66.40} \\ \hline
\end{tabular}
\caption{Comparison of our models with the defined baselines. We can observe that our model reports the best performance compared to the other solutions.}
\label{tab:results}
\end{table}

\begin{figure}
    \centering
    \includegraphics[scale=0.57]{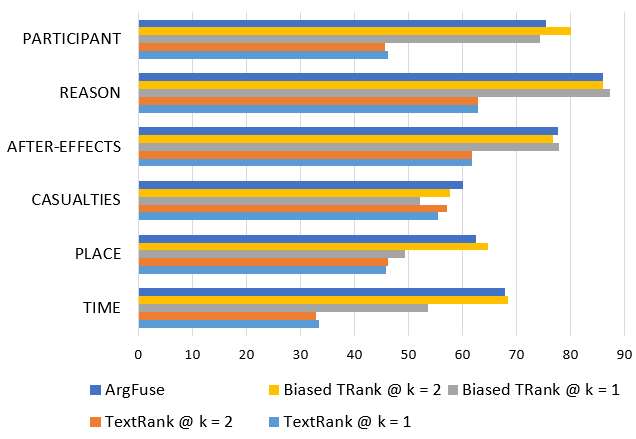}
    \caption{Comparison of Argument-Wise F1-Scores.}
    \label{fig:arg_scores}
\end{figure}

\begin{figure*}[t]
    \centering
    \includegraphics[scale=0.7]{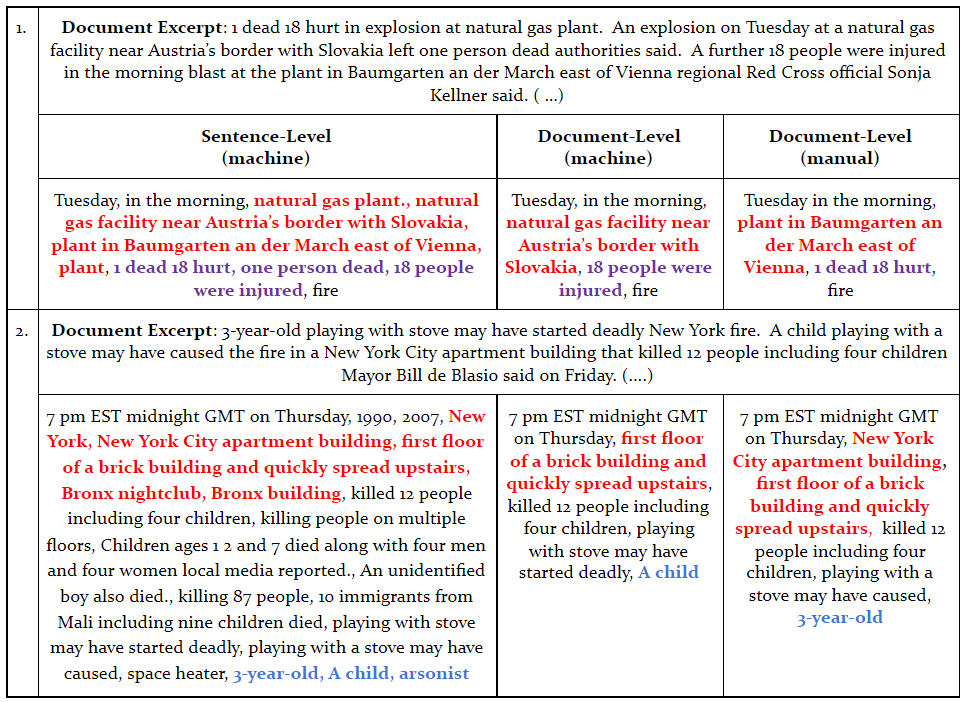}
    \caption{Comparison of sentence-level outputs with reduced document-level machine and human outputs. Highlighted phrases refer to the arguments that differ from the document-level gold standard. Phrases in red are the \textit{Place} arguments while the ones in Purple and Blue are the \textit{Casualties} and the \textit{Participant} arguments respectively.}
    \label{fig:example}
\end{figure*}

\subsection{Results}
We present our findings and results for each module in Table \ref{tab:module-wise}. For evaluating the results of our complete framework, we first prepare our reference text to which the machine output will be compared. Each of the manually curated information frames are presented as a sentence in the reference text in a newline. The presented  sentence contains the aggregated arguments for each argument type separated by the comma delimiter. We compare our final framework with three baseline models:
\begin{itemize}
    \item GiveMe5W1H \cite{Hamborg2019b}: is an unsupervised approach for extracting document level phrases related to the six 5W1H questions (what, where, when, who, why, and how) from English News Articles. We map the six questions to our argument types so that we could run the system on our dataset and evaluate.
    \item TextRank \cite{mihalcea-tarau-2004-textrank}: We use the graph based ranking algorithm to rank the sentence-level argument phrases for each argument type in a document extracted using \cite{kar2021event}'s model. We select the top-k arguments as the representative arguments for that argument type from the document.
    \item Biased TRank: Similar to the above described baseline, but instead of TextRank we have used Biased TextRank \cite{kazemi-etal-2020-biased} to rank the extracted sentence level arguments directly.
\end{itemize}
The final results are reported in Table \ref{tab:results}.

\section{Analysis}
\begin{figure*}
    \centering
    \includegraphics[scale=0.7]{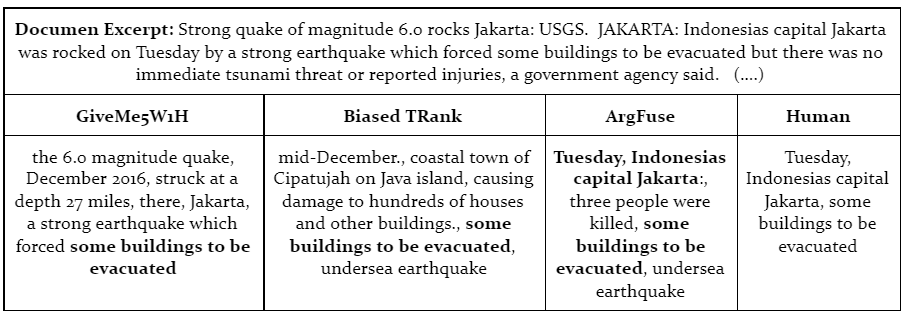}
    \caption{Comparison among various baseline models and ArgFuse. The highlighted arguments are the ones that match with the gold standard output.}
    \label{fig:compare}
\end{figure*}

In this section, we present a thorough analysis of our findings and the novel task that we introduce. We investigate the contributions as well as the pitfalls of our framework and attempt to provide directions for improvement. In the sections to follow, we analyse our framework both quantitatively and qualitatively.

\subsection{Quantitative Analysis}
In this section we shall scrutinize the performance of each module in our framework as well as analyze the overall performance. The module wise performance is reported in Table \ref{tab:module-wise}. The final results are reported in Table \ref{tab:results}. We can observe that while the completely unsupervised method (GiveMe5W1H) has a poorer extraction capacity, the other semi-supervised models exhibit a stronger performance. We also observe that the approaches processing contextual information like Biased TRank and our model ArgFuse report a much higher performance thus highlighting the importance of contextual information in this task. We find that while Biased TRank @ k = 1 reports the highest precision score, Biased TRank @ k = 2 reports the highest recall score. The stark difference between the precision and recall values of these two baseline methods is reflective of the problem of fixing a suitable 'k' value for the task of aggregation. While the higher number of argument roles with a single mention in our dataset (\textit{singles} as illustrated in Figure \ref{fig:dataset2}) is favourable for a higher precision value for Biased TRank @ k = 1, allowing the inclusion of more arguments in the final document frame can be attributed to the high recall score of Biased TRank @ k = 2. However, fixing the number of argument mentions to be included in the final document frame increases the count of false negatives for k = 1 and increases the number of false positives for k = 2. Our model presents a dynamic approach of including any number of relevant and precise arguments for representing document-level information. ArgFuse reports state-of-the-art results with a an acceptable balance between the precision and recall values.\\
In Figure \ref{fig:arg_scores}, the argument-wise results of the various baseline models along with that of ArgFuse is presented. We can observe that for most of the arguments, ArgFuse and Biased TRank report the highest performance, with ArgFuse reporting better or comparable performance in 4 out the 6 arguments. We find that ArgFuse reports comparatively poorer performance for arguments like \textit{Reason} and \textit{Participant} which constitute the argument classes with least number of samples (as depicted in Figure \ref{fig:dataset1}). Also, argument classes like \textit{Reason} mostly constitute of \textit{singles} as depicted in Figure \ref{fig:dataset2} and hence reports a higher score using Biased TRank @ k = 1. 

\subsection{Qualitative Analysis}
In this section we closely look at the failure cases for error analysis.
In Figure \ref{fig:example}, we record two examples from our dataset with both human and machine annotated information frames which are representative of the generic merits and demerits of our model. In the first  example, we notice that instead of selecting ``Baumgarten'', the mention containing the NERs ``Austria'' and ``Slovakia'' is chosen. This is indicative of the bias of the ranking process towards more popular named entities and might not favour mentions containing rare named entities. In the second example, for the \textit{Place} arguments, we observe that ``New York, New York City apartment building'' are rightly deemed redundant by our redundancy classifier, and the argument ``New York'' is discarded. When the arguments ``New York City apartment building, first floor of a brick building and quickly spread upstairs'' are evaluated, they are incorrectly classified as redundant. When the argument mentions, such as the above mentioned pair, are very similar in both their surface forms and content, the model sometimes fails to capture their exclusivity in terms of information.
In case of the participant arguments, we observe in the second example that the irrelevant argument ``arsonist'' is correctly discarded. However, for both the \textit{Casualty} arguments in the first example and the \textit{Participant} arguments in the second example, the relevant arguments are again very similar in nature, and the model ranks the incorrect argument mention over the correct ones. For example, in example 1, ``18 people were injured'' is ranked higher than the other candidate argument mentions thus resulting in some loss of information.

In Figure \ref{fig:compare}, we present the comparison between the outputs generated by ArgFuse and some of the other baseline models. We observe that the outputs retrieved from both GiveMe5W1H and Biased TRank present with irrelevant and redundant content. The output of ArgFuse correlates the most with the gold standard output for the document. This can be regarded to the explicit relevance and redundancy checks in the ArgFuse algorithm which mines precise document level information frames effectively.

\section{Conclusion}
In this paper, we present an extractive approach to aggregate sentence-level argument or entity mentions to produce precise document-level information frames from lengthy text articles effectively. With a very scarce amount of work being conducted in the field of document-level IE, we develop and open-source our dataset of aggregated argument mentions. To the best of our knowledge, we present the baseline for the task of argument aggregation and open-source our work for research. We closely analyse the merits and demerits of the model and encourage scientists to build on the pitfalls discussed and enhance the aggregation capabilities at a document level. For future work, we want to analyse the model's aggregation capabilities in crosslingual and multilingual environments and extend the aggregation capabilities across document boundaries. As explored in works like \cite{10.5555/1565754.1565782, ji-grishman-2008-refining}, extending \textit{ArgFuse} to aggregate information from multiple news sources can present with a very useful and practical use case. 

\section*{Acknowledgments}
The work done in this paper is an outcome of the project titled “A Platform for Cross-lingual and Multi-lingual Event Monitoring in Indian Languages”, supported by IMPRINT-1, MHRD, Govt. of India, and MeiTY, Govt. of India.


\bibliographystyle{acl_natbib}
\bibliography{anthology,acl2021}


\end{document}